\documentclass[journal,12pt]{IEEEtran}
\usepackage{graphicx}
\usepackage{amsmath,amssymb} 
\usepackage{color}
\usepackage{layout}\textwidth = 7in
\usepackage[author={Max Schlepzig}]{pdfcomment}
\usepackage{amsmath}
\usepackage{color}
\usepackage{times}
\usepackage{epsfig}
\usepackage{graphicx}
\usepackage{amsmath,bm}
\usepackage{amssymb}
\usepackage{hyperref}       
\usepackage{url}            
\usepackage{booktabs}       
\usepackage{amsfonts}       
\usepackage{nicefrac}       
\usepackage{microtype}      
\hypersetup{colorlinks=true, linkcolor=red}
\usepackage{amsmath}
\usepackage{cite}
\usepackage{mathrsfs}
\usepackage{soul}
\usepackage{multirow}
\usepackage{comment}
\usepackage{rotating}
\usepackage{amssymb}
\usepackage{pifont}
\usepackage[sort&compress,square,comma,numbers]{natbib}
\newcommand{\etal}{\textit{et al}. }

\begin{document}
\title{ A Survey on Multi-Objective Neural Architecture Search}
\author{Seyed Mahdi Shariatzadeh,
        Mahmood~Fathy,
                Reza~Berangi,
        and~Mohammad~Shahverdy
\thanks{Department of Computer Engineering, IRAN University of Science and Technology(IUST), Tehran, IRAN. E-mail:{m\_shariatzadeh@comp.iust.ac.ir, mahfathy@iust.ac.ir}.(Corresponding author: Mahmood Fathy)}
}

\maketitle
\begin{abstract}
Recently, the expert-crafted neural architectures is increasing overtaken by the utilization of neural architecture search (NAS) and automatic generation (and tuning) of network structures which has a close relation to the Hyperparameter Optimization and Auto Machine Learning (AutoML). After the earlier NAS attempts to optimize only the prediction accuracy, \textit{Multi-Objective Neural architecture Search (MONAS)} has been attracting attentions which considers more goals such as computational complexity, power consumption, and size of the network for optimization, reaching a trade-off between the accuracy and other features like the computational cost. In this paper, we present an overview of principal and state-of-the-art works in the field of MONAS. Starting from a well-categorized taxonomy and formulation for the NAS, we address and correct some miscategorizations in previous surveys of the NAS field. We also provide a list of all known objectives used and add a number of new ones and elaborate their specifications. We have provides analyses about the most important objectives and shown that the stochastic properties of some the them should be differed from deterministic ones in the multi-objective optimization procedure of NAS. We finalize this paper with a number of future directions and topics in the field of MONAS.
\end{abstract}


\section{Introduction}
Architectural engineering is an essential need for deep learning, that is often done by experts. This is an error prone and time consuming task; so the Neural Architecture Search has been employed to automatically find a suitable architecture according to certain objectives. NAS is a subfield of automated machine learning and is related to hyperparameter optimization \cite{AutoMLsurvey}.

Several review articles on the search for neural architecture have been written so far, but none is dedicated on the multi-objective NAS. \cite{nassurvey} is an early review paper on NAS which covers the overall concepts of NAS without dealing it as an optimization aspect. \cite{asurveyonnas} contains a detailed review of commonly adopted architecture search spaces and search methods. \cite{acomprehensivesurvey} presents an overview of the earliest NAS algorithms and their problems. In \cite{nassurvey2} a recent review on the NAS and hardware accelerator co-search is provided which categorizes the concepts in three classes of Single-Objective NAS, Hardware-Aware NAS, and NAS with hardware co-optimization. \cite{AutoMLsurvey} provides a comprehensive review of state-of-the-art in automated machine learning (AutoML), covering the Data Preparation,  Feature Extraction, Hyperparameter Optimization, and the NAS and provides a good description of the primary methods of NAS. In
\cite{benmeziane2021hardware} a new taxonomy of Hardware-Aware NAS is provided and there, the hardware-aware NAS is defined such a multi-objective optimization problem. 

Compared to previous review articles, our article is devoted to MONAS. We have also tried to provide a more accurate formal model of the problem. Following our notation, our classification of NAS methods slightly differs from all the previous surveys and taking advantage of a stronger logic, corrects them. In the Architecture Search section, we will present analyzes that show the effect of the stochastic nature on training and inference of samples in real neural networks, which will have a serious impact on network evaluation, a topic that is always present in NAS but is generally ignored. We also provide a complete list of types of objectives that can be used for MONAS, which is more complete than all known references and also contains several novel objectives.

We will continue this paper as follows: First, we will have an overview of the concepts of neural architecture search, so that we can define the general definitions in a well-defined and accurate way. A formal definition of the problem is provided in this section. In Section III, we discuss the topics of multi-objective neural architecture search, including Pareto optimization, vectorization, and the types of targets that can be used for MONAS. In Section IV, we review the most important previous works of MONAS, and in various dimensions, we briefly review the key articles. Section V further addresses executive issues. We conclude this paper with a summary and reference to the topics ahead in MONAS.

\subsection{Neural Architecture Search}
The need for NAS arises because the design and optimization of the deep neural network has several architectural parameters including the properties of each layer such as the number and type of operators, and the number and how the building-blocks are connected to each other. The choice of hyperparameters and architecture for a particular application can increase or decrease specifications such as accuracy, network size, computational cost, the amount of hardware required for training and inference, and so on. By automating the architecture determination process and estimating architectural parameters, the final neural network(s) can be specified automatically through a process of repeated optimization or improvement. In this case, by having an efficient search space and using the appropriate primary elements, better architectures can be identified by spending an appropriate amount of time.

\begin{figure*}[htbp]
\begin{center}
\includegraphics[width=.85\linewidth]{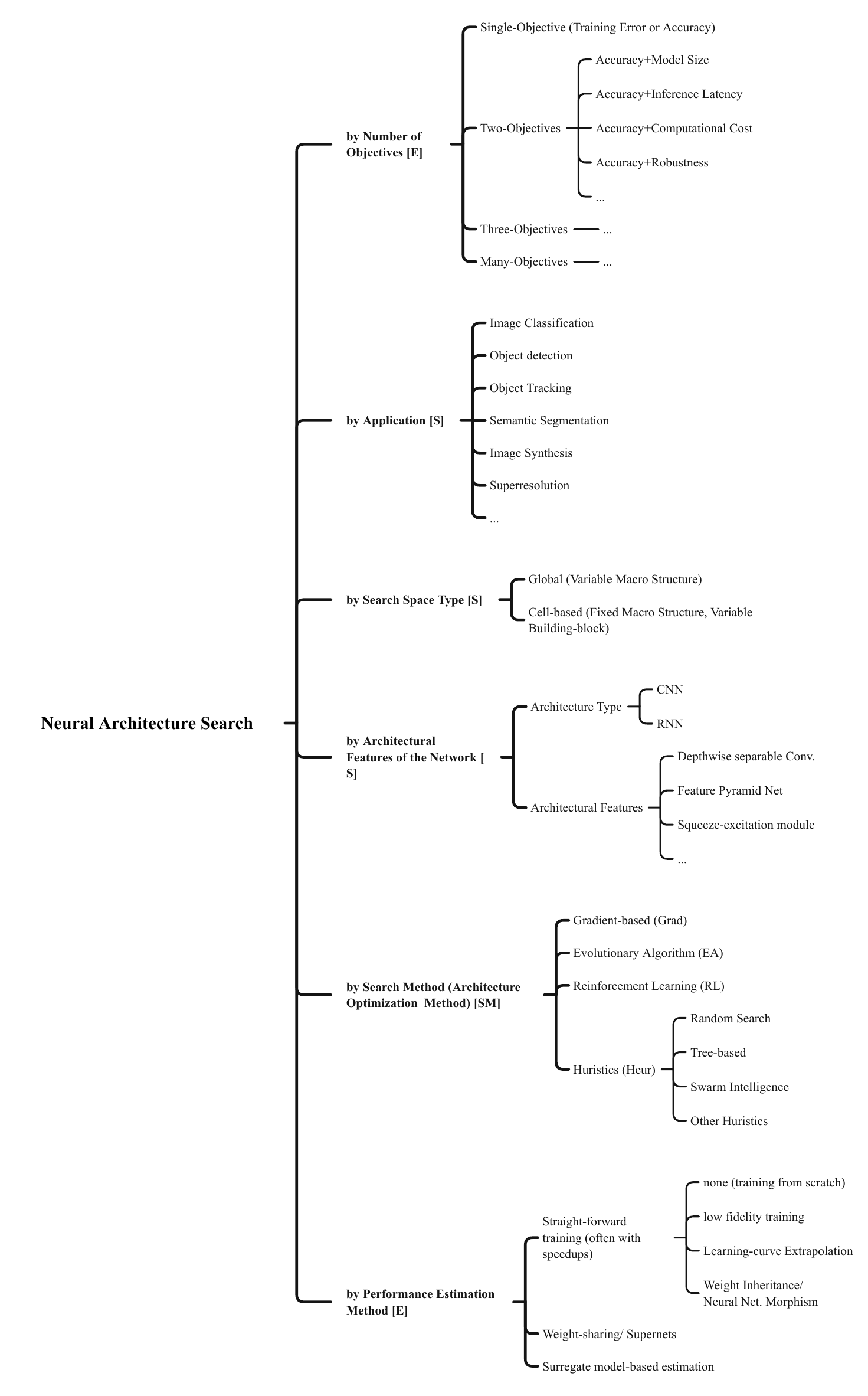}
\end{center}
 \caption{A taxonomy for NAS; the related component in the NAS procedure is marked in brakets: [S] indicates Search Space, [SM] is for Search Method, and [E] for Performance Estimation.}
\label{fig:taxonomy}
\end{figure*}

\section{Formal Problem Definition}
In order to maintain the accuracy in our study, an accurate notation for defining the NAS as well as the training procedure is needed, which we provide here.

As training a DNN is itself an optimization problem (often done by algorithms such as the Stochastic Gradient Descent), at NAS we actually have a bilevel optimization problem: on the one hand, for a certain architecture, we have to achieve efficient values for the weights and biases, an optimization (network training) should be done, and on the other hand, the architecture itself should be optimized.

We define a neural network $N$ as a pair of its \textit{Architecture} $\alpha$ and \textit{parameters} (or \textit{weights}) $w$. \cite{kandasamy2018neural} and \cite{talbi2020optimization} also proposed a good notation for DNNs that has similarities to ours.
\begin{equation}
N=(\alpha, w).
\label{eq:n}
\end{equation}

The network architecture $\alpha$ is a pair of the \textit{network topology graph} and \textit{layer definitions}. The network topology is a directed graph $\mathcal{G}=(L, E)$ (a directed acyclic graph in CNNs) with nodes $l \in L$ as network layers and represents the operations and their order in the neural network:

\begin{equation}
\alpha=(\mathcal{G}, F),    \mathcal{G}=(L, E).
\label{eq:alpha}
\end{equation}

Each $l$ in $L=\{l: 1 \leqslant l \leqslant l_{max}\}$ corresponds to a layer in the neural network which transforms its input data by a function $f^{(l)}$. The function $F: L \rightarrow \Omega_F$ defines the mathematical operation of each layer $l$, as $\Omega_F$ represents the set of all possible layer functions in the neural networks. The network parameters $w$ is a function that provides all the required parameters for each layer, as $w: L \rightarrow \Omega_{w}(F)$.

We illustrate the whole mathematical operations of a neural network as a neuro-calculation function $f(.)$ which calculates the output prediction $p$ of the network  for the network architecture $\alpha$ and network weights $w$ for an input data $x=x_d$ from a sample data $d=(x_d, y_d)$ \cite{Goodfellow-et-al-2016}:
\begin{equation}
p=f_{\alpha, w}(x).
\label{eq:f}
\end{equation}

The network response function $f(.)$ is made from a number of data manipulation layers such as convolution, local and global pooling, batch normalization and so on, each of which performs mathematical operation $f^{(l)}(.)$ on its input $I^{(l)}$ and produces the resulting $O^{(l)}=f^{(l)}(I^{(l)})$ for the layer $l$. 

%

We can define the NAS as an optimization problem: For a desired cost function $cost(., .)$ which rates the cost of a trained neural network for a given validation set, we are searching for the best architecture(s) $\alpha^*$ which minimizes the aforementioned cost function over a subset of the space of all the possible architectures $S_{\alpha} \subset \Omega_{\alpha}$:

  \newcommand{\argmax}[1]{\underset{#1}{\operatorname{argmax}}\;}
  \newcommand{\argmin}[1]{\underset{#1}{\operatorname{argmin}}\;}

\begin{equation}
\alpha^*=\argmin{\alpha \in S_{\alpha} \subset \Omega_{\alpha}} cost(f_{\alpha, w^*}, D_{val}),
\label{eq:nas}
\end{equation}

\begin{equation}
w^* = \argmin{w\in \Omega_{w}}Loss(f_{\alpha, w}, D_{train}).
\label{eq:b}
\end{equation}

\begin{equation}
\begin{split}
Loss(f_{\alpha, w}, D_{train}) = \sum_{(x_d, y_d) \in D_{train}}{loss(y_d, p)}\\
=\sum_{(x_d, y_d) \in D_{train}}{loss(y_d, f_{\alpha, w}(x_d))}.
\end{split}
\label{eq:c}
\end{equation}

In the above equations $w^*$ represents the best (found) network parameters during the learning procedure which minimize network loss on the whole training set. Performing the minimization of \autoref{eq:b} is known as \textit{training}. $loss(., .)$ is the loss function used in training the network (such as the Cross-Entropy or Mean Squared Error) and $y_d$ represents the label of training sample data $d=(x_d, y_d)$ from the training set $D{train}=\{d^{(i)}=(x^{(i)}, y^{(i)})\}$. $L(.,.)$ is the total loss value of a network on a given dataset.

In practice, instead of searching among all the possible architectures, we search in a limited cases that always obey a certain paradigm. As an example, the search in \cite{efficientnet} is to find optimal values for network width, depth and resolution; therefore there is only three degrees of freedom for architecture search and the search is done in $(w, d, r) \in \mathbb{N}^3$ space. So we can define a more restricted search space for the optimization as the \textit{architecture hyperparameter search space} ($S_h$) and define the NAS equation for the architecture hyperparameters. A \textit{Generator Function} can produce the architecture $\alpha$ from the architecture hyperparameters $h$ as:

\begin{equation}
\alpha = G(h).
\label{eq:gen}
\end{equation}
The formulation of the NAS optimization (\autoref{eq:nas}) can be rewritten as follows:

\begin{equation}
\alpha^*=G(\argmin{h \in S} cost(f_{G(h), w^*}, D_{val})).
\label{eq:nas2}
\end{equation}

In the trivial form $G(h) = h, S = S_{\alpha}$, so the new form of the NAS Optimization does not restrict the definition.

The above yields to a slightly different definition for NAS: We are searching in the architecture hyperparameter space for architecture hyperparameters that reduce the cost function. We will refer to the \autoref{eq:nas2} as the \textit{NAS Optimization} and \autoref{eq:b} as the \textit{Training Optimization}. \autoref{tab:elems} provides a list of all of our definition.

Note that the above NAS Optimization formulation has not any specifications of the neural networks and describes a wider area of algorithm search in automated machine learning (AutoML). Also, it is not always necessary to perform the training optimization first and then calculate the NAS optimization and the parts of the bilevel optimization can be performed together (as done in differentiable NAS methods).

\begin{figure*}[]
\begin{center}
\includegraphics[width=.5\linewidth]{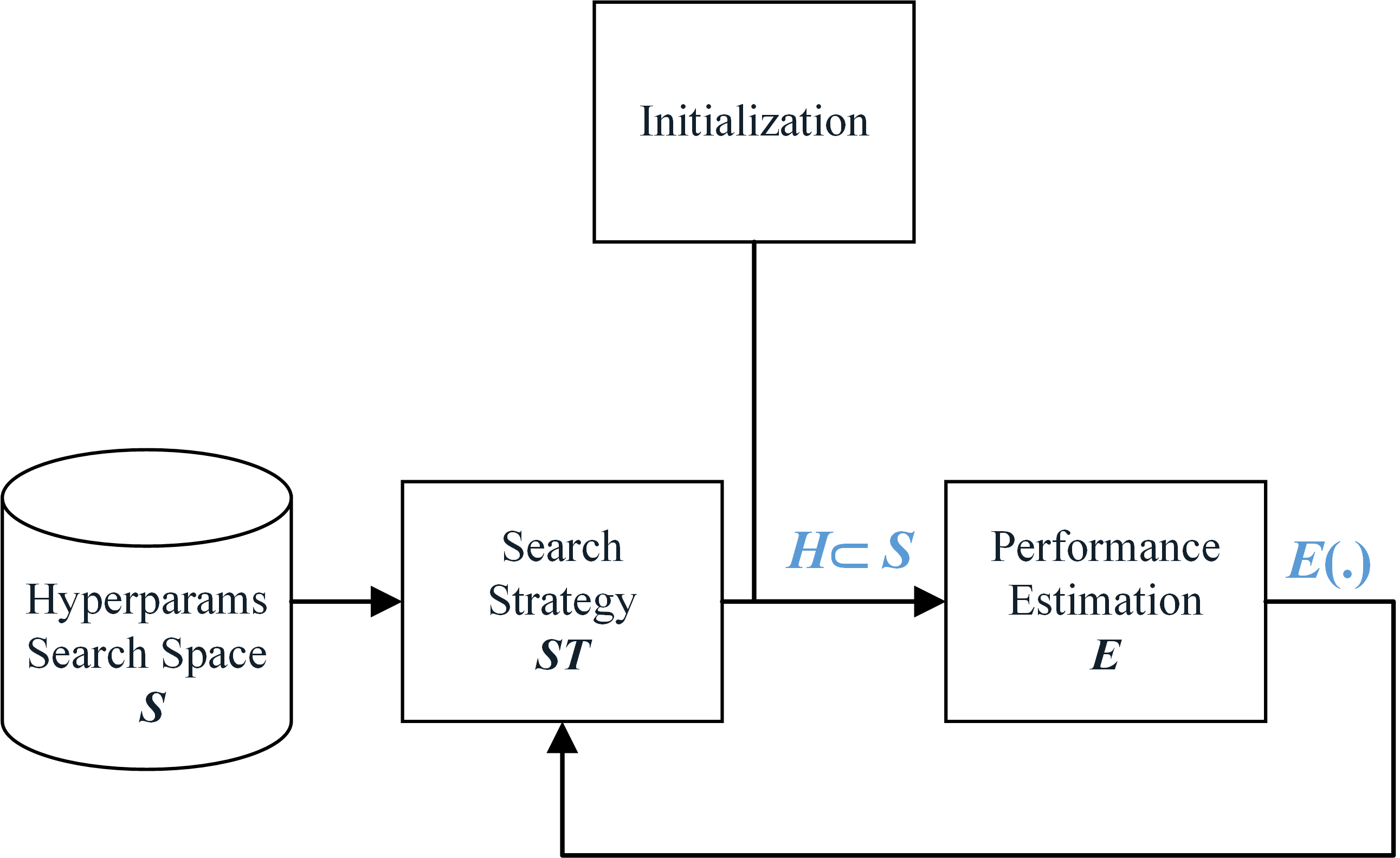}
\end{center}
 \caption{Necessary Components of a NAS. The Search Space is the definition of the NAS space and two main stages of Search Method and Performance Estimation are connected in a feed-back configuration.}
\label{fig:nasdiag}
\end{figure*}

\begin{table}[]
\caption{List of Principal Elements of a NAS operation (See text for details).}
\label{tab:elems}
\begin{center}
\begin{tabular}{l l l l }
\hline
Component   &   Symbol  \\
\hline
Architecture Search Space & $S$ & \\
Search Method & $SM$ & \\
Initial Hyperparams & $H^0$ & \\
Candidate Hyperparams & $H$ & \\
Architecture Generator & $G(H)$ & \\
Candidate Architectures & $\alpha=G(H)$ & \\
NAS Datasets &  $D_{train}, D_{val}$ &  \\
Target Datasets &  $D'_{train}, D'_{val}$ &  \\
Performance estimation & $E(\alpha)$ & \\

Termination Criteria &  $T$ &   \\
Final Architecture(s) & $\alpha^*$ & \\

\hline
\end{tabular}
\end{center}
\end{table}

\section{Principal Elements of NAS}
For any NAS experiment three main components must be determined: the Search Space, the Search Method (i.e. the Optimization Method \cite{nassurvey2} or Architecture Optimization \cite{AutoMLsurvey}), and the Cost Estimation (or Performance Estimation \cite{nassurvey}) Strategy. In other words, in \autoref{eq:nas2}, the term $S_{\alpha} \subset  \Omega_{\alpha}$ illustrates the Search Space, the $\argmin{.}$ relates to the Architecture Optimization, and $ cost(f_{G(h), w^*}, D_{val})$ represents the Cost Estimation. \autoref{fig:nasdiag} illustrates the relation of the overall parts of a NAS. We also added an Initialization step to this diagram to make it complete, which will be described  together with the Search Method. As can be seen in \autoref{fig:nasdiag}, the Search Method (in the first step, the Initialization method) provides a set of one or more architectural hyperparameters $H=\{h^{(i)}\}$ from the search space which should be evaluated by the Performance Estimation routine. The result is fed back to the Search Method to repeat the search mechanism. The same procedure is also performed in hyperparameter optimization methods in AutoML.

\begin{figure*}[htbp!]
\begin{center}
\includegraphics[width=.75\linewidth]{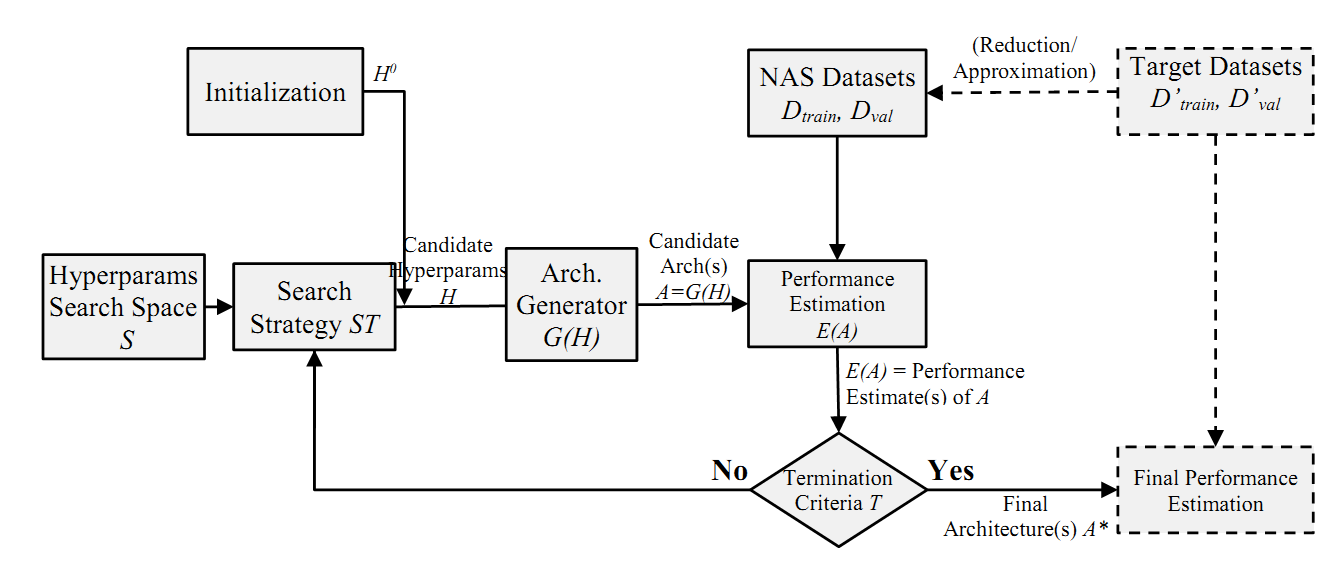}
\end{center}
 \caption{A detailed view of components of NAS experiments including postprocessings. The Architecture Generation is usualy performed implicitly.
}
\label{fig:nasdiagram}
\end{figure*}

A more elaborated overal procedure of NAS can be viewed in \autoref{fig:nasdiagram}:
From the initilizing architecture hyperparameters $H$ which relates to a network architecture $G(H)$, performance estimations are provided which may include error rate, size, latency and so on. New guesses for hyperparameters that produce networks with better performances are provided according to the search strategy. At some termination criteria based on maximum NAS steps, performance reached, overall procedure time and so on, the feed back loop of the optimization will be terminated. The whole above procedure may be run on a smaller dataset and at this time, the found architecture may be altered (e.g. made deeper or wider) and retrained on a (larger) target dataset. Note that the architecture generation is a logical operation that may implicitly done. In some methods (such as one-shot NAS) the network training operation (using backpropagation) is performed in the initialization, otherwise it is performed in the performance estimation stage.

In this section, we refer only to the generalities of the NAS discussion to the extent that the most important parts of it are introduced, and we will omit the details to have more focus on the multi-objective NAS. For a more complete discussion of general NAS (mostly for single-objective methods), see \cite{asurveyonnas, acomprehensivesurvey, nassurvey, nassurvey2}.

As mentioned earlier, the three main components of NAS are:

\subsection{Search space}
The architecture search space $S_{\alpha}$ is the set of all the different architectures that are possible solutions of the Architecture Optimization, which should be considered and searched to find the best architecture. The set of all representable architectures $\Omega_{\alpha}$ may contain many architectures that are not useful, such as architectures that perform unsed computations or are even errorneous, depending on the specifications of the description used for neural networks. Also, the search space is almost always restricted to a certain class of neural networks with a restricted set of operations with an almost fixed structure. For example, architectures with several cascaded pooling layers mostly have poor performances and is out of the search space of most of the prior art, eventhough there are corresponding architectures in $\Omega_{\alpha}$. These restrictions are applied implicitly in the representation used and differs from the policies applied by the search algorithm.

As mentioned before, in practice the optimization is done on few architectural parameters that can be referred to as $h$ which belongs to the corresponding space $\Omega_{h}$ (named as \textit{Encoding Space} in \cite{liu2021survey}). We differ two search spaces $S_{\alpha}$ and $S$ while previous surveys on NAS \cite{asurveyonnas, acomprehensivesurvey,nassurvey,nassurvey2,AutoMLsurvey} didn't.

We categorize NAS search spaces into two types:

\subsubsection{Variable Macro Structure (also known as global search space)}
In this category that was used in all the earlier works on NAS, the overal structure of the network may be altered and the search may change the macro properties such as the number of layers \cite{baker2016designing, suganuma2017genetic}. Global search space may be defined to allow very large changes in the network or may be restricted to only a few (as 3) architecture hyperparameters (such as depth, width, resolution), as in EfficientNet \cite{efficientnet}.
Note that the speed of a NAS method is not related to the level of restriction of the search space. As an example, in ENAS \cite{enas}, a much bigger search space is used while the overal procedure is faster than \cite{efficientnet} with only 3 architecture hyperparameters.

\subsubsection{Fixed Macro Structure, Variable Building-block structure (aka Cell-based Search Space)}
 This model focuses only on manipulating the structure of restricted building-blocks that will shape the whole network while all the other features of the network (parameters like the number of network layers, filter sizes, and so on) are fixed and is also known as \textit{inner-level optimization}\cite{talbi2020optimization}. This method is motivated by the fact that optimal building-blocks can be easily transfered to larger problems, as applying the transfer learning paradigm on the NAS. NASNet \cite{nasnet} was an early work in this domain that achieved results better than handcrafted architectures for image classification problems by searching on the CIFAR-10 dataset and using the resulting blocks for Imagenet classification. NASNet uses 13 choices for operation layers (such as 3x3 convolution and 5x5 max pooling) in a search for finding two efficient distinct building-blocks named \textit{normal cells} and \textit{reduction cells}. Many cell-based search spaces only allow altering the connection between operands in the building-blocks while keeping every other properties of the network fixed \cite{darts}. Similar search space strategy is used by other optimization methods \cite{BlockQNN,pnas,hierarchicalNas, darts, pdarts}.

It worths noting it is shown that if the search space is defined efficient, even the random search strategy can produce competiting DNNs \cite{xie2019exploring, li2020random}.

\subsection{Search Method (Optimization Method)} 
The main task of the NAS is the architecture search, as the other stage (performance evaluation) is performed in just training any DNNs. Back to the NAS optimization formulation (\autoref{eq:nas2}), we need to perform an optimization on the $cost(.,D_{val})$ function which may be an expensive operation. The search space is usualy an infinite space mostly containing useless answers. The search method should propose good possible answers according to recent performance estimations (which is done using optimization methods such as genetic algorithm). In the following we provide a brief introduction to search methods for NAS.

\begin{table*}[]
\caption{The corresponding concept for each category of NAS methods in the general NAS model (\autoref{fig:nasdiag}).}
\label{tab:stages}
\begin{center}
\scalebox{.85}{
\begin{tabular}{l l l l l }
\hline
Stage/ Concept  & Grad	&   RL & Evo	& Heur	 \\
\hline
The set of hyperparams	& random initial probabilities 	& network encodings	& EA population	& any	 \\
Initialization	& - & - 	& generate the first population	& -	\\
Search Method & model and fine-tune a candidate & rl	& one EA generation & a specific heuristic method	\\
Performance Estimation	& & \\
Finalization	& produce and train the final architecture & return the final architecture(s)	& return the final architecture(s)	& return the final architecture(s)	\\
\hline
\end{tabular}
}
\end{center}
\end{table*}

\subsubsection{Evolutionary Algorithms (EA)}
Evolutionary algorithms (such as genetic or swarm intelligence methods like ant colony algorithms) provide solutions for stochastic global optimization of black-box functions by means of an evolving population of encodings of potential solutions (individuals) for a number of generations.

A typical genetic evolutionary algorithm consists of initialization, crossover (recombination), mutation, and 
selection.
EA can be used to optimize architectures as well as training hyperparameters for Neural Networks \cite{angeline1994evolutionary}.
The early work of 'NeuroEvolution of Augmenting Topologies (NEAT)' \cite{neat} provided a genetic encoding for neural networks which further extended to CoDeepNEAT for DNNs in \cite{miikkulainen2019evolving} by presenting each layer of DNN by a chromosome rather that a single node and evolving monotonous modular structures proficiently. The network structure is assembled by a network blueprint in which, each node is replaced by a module which is a structure in many successful recent DNNs. More global hyperparameters (such as image data augmentation parameters) are also used.
EA-NAS is utilized for image classification and image captioning, among the other tasks \cite{miikkulainen2019evolving, real2019regularized, ijcai2018, real2017large, galvan2021neuroevolution}.

In \cite{real2019regularized}, authors evolve an image classifier which is called AmoebaNet-A by evolutionary techniques. The accuracy of AmoebaNet-A is comparable with ImageNet models. However, the computational cost of the search process is very high, approximately 3150 GPU-days. The authors show experimentally that EA is better search method in the limited resource situations, because EA finds more accurate models in the early generations in comparison with other approaches.

\cite{kandasamy2018neural} proposed \textit{'NAS with Bayesian Optimisation and Optimal Transport'} (NASBOT), in which the genetic algorithm is used as the search strategy and candidate model generation in a statistical and Bayesian Optimization context.

In \cite{liu2021survey} a survey on evolutionary algorithms for NAS is provided. It also contains a brief description of multi-objective evolutionary algorithm and swarm intelligence NAS experiences.

\newcommand{\multiobjective}{\textit{Multi-objective Notes:}}
\multiobjective DeepMaker \cite{loni2020deepmaker} is a framework for automatic design of DNNs on limited processing units like sensors. It follows a multi objective evolutionary approach for optimizing accuracy and network size. NSGA-net \cite{lu2019nsga} is another multi-objective evolutionary approach for NAS. Its objectives are accuracy and computational complexity measured by FLOPs. In \cite{elsken2018efficient} a multi-objective evolutionary algorithm is used for optimizing both prediction accuracy on 2 datasets and complexity in terms of number of parameters or number of multiply-add operations or inference time. 
The proposed approach, LEMONADE, also controls enormous computational resources needed by EA. In order to reduce the computational resources, child networks are not trained from scratch, and they inherits their parent's 
learned parameters and training abilities. 
Moreover, to generate children from parents, first, parent population is sampled based on cheap objectives like number of parameters and then based on expensive objectives like network accuracy. This strategy also decreases the computational resources needed by EA. This paper will again refer to multi objective evolutionary approaches in \autoref{sec:monas}, when it discuss multi objective approaches in general.

\cite{cai2020once} used the evolutionary algorithm of \cite{real2019regularized} based on the neural-network-twins for performance evaluation with accuracy and (embedded) hardware inference latency.

\subsubsection{Reinforcement Learning}
Reinforcement Learning (RL) \cite{sutton1998reinforcement} methods are the pioneering methods for the architecture search when a high computational resources are available. Generally, RL methods model an agent that, at each step $t$, tries to maximize the reward value $r^{(t)}$ in interactions with an environment by taking action $a^{(t)}$. For NAS, the search algorithm acts as the agent that tries to maximize the performance of its generated architecture according to the performance estimation measure \cite{baker2016designing}.

A successful method for applying RL for NAS is to train a recurrent neural network (RNN) (named as \textit{controller}) that returns a description (encoding) of the output DNN layer-by-layer (e.g. the controller gets the encoding of the last layer and produces the encoding of the next layer). The encoding consists of properties such as operation type, filter size and filter stride size). The controller is trained by the reward signal which comes from the environment which in fact is a training and evaluation procedure for DNNs on a validation set. A policy gradient method such as REINFORCE can be used to improve the efficiency of generated models \cite{zoph2016neural}. Other notable RL-NAS works include \cite{enas, nasnet, BlockQNN}.

The RL methods for NAS usually took a huge amount of computational cost. As an example, NASNet \cite{nasnet} took 2,000 GPU-Days of Nvidia P100 for an experiment on CIFAR-10 dataset. In ENAS \cite{enas} took advantage of sharing parameters among child models instead of training each one from scratch to achieve a 50000x speedup and perform the entire procedure in about 7 hours on a single GPU.

A review on Reinforcement learning techniques for NAS is provided in \cite{rlnassurvey}.

\multiobjective
In \cite{gong2019mixed}, both NAS and model quantization are used for selecting a hardware efficient DNN. 
The objective function is a weighted sum of loss function for a sample classification problem and deviation from a predefined energy threshold. In order to find a hardware efficient model, reinforcement learning is used and 
gradient decent is used for optimizing loss function. It is indeed the combination of RL and gradient decent approaches. The optimal precision choices (quantization bit-widths) are also considered as one of network architecture parameters.

MNASNet is a platform-aware neural architecture search approach for mobile that is proposed in \cite{tan2019mnasnet}. It looks for a good trade of between accuracy and latency where the latency is the response time of model on a Pixel phone. The objective function of MNASNet is define by a customized weighted product of the accuracy and the latency and it is optimised by RL. The search space of MNASNet preserves layer diversity in different blocks which construct the whole network. In \cite{howard2019searching}, the authors use a multi objective approach similar to \cite{tan2019mnasnet} and search for MobileNetV3 both for high and low resource use cases in order to increase the accuracy and decrease the latency of MobileNetV2. This paper uses a combination of efficient mobile building blocks to achieve its goals.

An accuracy-and-performance-aware NAS (APNAS) method is proposed in \cite{achararit2020apnas}. It is a multi-objective method in which performance is an estimation of cycle count, so there is no need of simulation on hardware devices to know the performance of the model. The objective function (reward function of RL algorithm) of APNAS is a weighed sum of validation accuracy and normalized cycle count of model. The authors proposed an approach for cycle count estimation based on the characteristics of the underlying hardware architecture and CNN model.

\subsubsection{Gradient-based search}
Supposing a differentiable search space and problem modelling, gradient-based methods (aka Continouous Search Strategey \cite{acomprehensivesurvey}) can be used for optimizing hyperparameters \cite{bengio2000gradient}. The differentiable architecture search methods \cite{darts, pdarts, ahmed2018maskconnect, cai2018proxylessnas} are instances of the gradient-based NAS which define and train a differentiable supernet. The usual method to describe a continuous search space is to consider every possible operations (including the identity function) and relax the categorical choice of a single operation using a Softmax function. After the convergence, the softmax will be approximated by only one operation (and connection) and the other ones are omitted in the finalizing stage. In this setup, training both connections and network parameters are performed simultanously in a single training procedure.

\textbf{The false category of One-shot NAS:} A One-shot NAS method is defined as an algorithm that trains only one neural architecture during a complete NAS experience, which is performed in the Initialization step. This follows the paradigm of eager algorithms. It worths noting that weight sharing methods such as \cite{enas, casale2019probabilistic} are sometimes categorized together with gradient-based methods as one-shot NAS (for example, in \cite{nassurvey}) while weight sharing is a method for fast performance evaluation rather than a search strategy for proposing new candidates. As an example, a random search strategy can be used with a weight-sharing performance evaluation. We consider weight-sharing as a complicated performance evaluation method. 
This is also correct for the concept of 'NAS by pruning' \cite{li2019partial, ding2022nap, zhang2021one, li2022pruning}.

\subsubsection{Random search}
Surprisingly, random search is a serious search strategy for NAS. It is shown that given a good search space definition, random search can generate competitive results \cite{nasnet, bender2018understanding}. As an example, the work \textit{'Understanding and simplifying one-shot architecture search'} \cite{bender2018understanding} simply uses random search as a search strategy to achieve state-of-the-art performance. In the NASNet \cite{nasnet} it is also reported by the authors that the reinforcement learning controller is only slightly better than random search and random search can also produce efficient results.

In \cite{bergstra2012random} it is shown empirically and theoretically that  for hyper-parameter optimization, randomly chosen trials are more efficient than trials on a grid; so there is no use to take the grid search heuristic into account.

\subsubsection{Other heuristics}

It is obvious that any other search or optimization strategy can be utilized in the NAS problem definition, namely the \textit{Monte Carlo Tree Search} (MCTS). In PNAS \cite{pnas}, authors proposed a progressive neural architecture search for learning the structure of CNN. They exploited a progressive decision process as Sequential Model-Based Optimization (SMBO) technique to increase the complexity of structure (from simple to complex) in order to learn a surrogate model quickly and improve the search space. Their method has more efficiency than the reinforcement learning method in \cite{zoph2016neural} in terms of number of models evaluated and total time of computation.

Even the local search can be used as the search heuristic for NAS, as reported in \cite{bosman2021local}.

Sequential model based optimization is used to search neural architecture search space in \cite{dong2018ppp}.

\subsection{Performance estimation} \label{sec:naspe}
Performance estimation is the stage of evaluating (or even predicting) the fitness and the value of the candidate models.
It is defined as the process of giving a rank to the performance of any specific architecture in search space when it is applied on unseen data. Note that for NAS, a performance ranking (rather that precise estimation) of the candidates can be sufficient.

As training and testing of all architectures is usually computationally too expensive, techniques to diminish the cost of performance estimation are very usual, such as Lower Fidelity estimates \cite{runge2018learning, real2019aging}, Learning Curve Extrapolation \cite{klein2016learning, baker2017accelerating}, Weight Inheritance/ Network Morphisms \cite{cai2018efficient, elsken2018efficient}, and Weight Sharing \cite{xie2018snas, darts}. Although it may seem at first that the practical efficiency estimation is very simple, but actually an important part in the literature belongs to the efficient and fast methods of efficiency estimation \cite{baker2017accelerating}.

We categorize the performance estimation methods as follows:

\subsubsection{Straight-forward training}
It is obvious that the simplest way for performance estimation in NAS is to train a full nerual network from scratch and return any required performance metrics such as network accuracy on a test set. For a small problem in which a full training doesn't take too long (or with enough hardware resources) this approach is practically applicable. 

\subsubsection{Straight-forward training with speedups}
Considering very large datasets and few computational resources, training many networks from scratch is hardly possible. Simple speed-ups can be used to accelerate the training procedure or estimate the final result, including reducing training set, and performance prediction based on few training epochs \cite{?}. 

\subsubsection{Weight-sharing and Supernet ($\sim$One-shot)}
Weight-sharing technique takes advantage of the specifications of a certain search space: when the search space only includes networks that are subsets of a single \textit{supernet}, it is possible to train the supernet once and, for evaluating a specific network, only cut the unused connections to achieve the trained subnetwork. In this manner, given a trained supernet, the performance estimation takes the time of only evaluating a network on a test set.

\cite{DBLP:journals/corr/abs-2104-14545b} used weight-sharing acceleration along with evolutionary search algorithm for object tracking and cought state-of-the-art performance with 13x speedup to hand-crafted networks.

\subsubsection{Surrogate Model-Based}
Surrogate Model-Based Optimization learns a model or surrogate function which can predict the performance of a structure without actually training it.

In \cite{pnas}, the same search space of NASNet is used and the performance is estimated using an LSTM (named \textit{the predictor}) which reads a given sequence of encoding of a network and predicts its accuracy on the training set. (Their search strategy is heuristic tree search which, starting from a simple trivial network, for a next choice, the predictor LSTM predicts every possible candidates and for the top-$K$ ($\sim$100) best scores, the real accuracy values are evaluated, the predictor is updated, and the candidates are passed to the next level, searching in a progressive manner.)

\cite{cai2020once} for performance evaluation used a 3-layer feed-forward neural-network-twins with 400 hidden units in each layer as the accuracy predictor. They concatenate architectural features vectors into a vector that represents the neural network architecture and input image size, which is then fed to the three-layer feed forward neural network to get the predicted accuracy. The accuracy predictor network is trained with 16.000 random architectures, each of which is evaluated with a fast supernet mechanism. The root-mean-square error (RMSE) between predicted accuracy and estimated accuracy on the test set is reported to be only 0.21\%. 

\subsection{Initialization and Finalization} 
Initialization and Finalization are not always necessary stages of a NAS procedure, but certain search methods or performance estimation speed-ups may require pre-or-post-stages (see \autoref{tab:stages}). 

Gradient-based search methods require a random initialization and after reaching the stopping criteria, give a solution with soft connections which should be pruned. The resulting architecture should be fine-tuned or trained from scratch to give the final network.

When taking advantage of early stopping or training on a subset of the training set, it is necessary to train again or at least, complete the training to reach the final network parameters. Also supernet method for fast performance estimation results in a network structure that has sub-optimal weights and should be trained from scratch for better performance.

\section{Multi-Objective Neural Architecture Search} \label{sec:monas}
A single-objective NAS tries to optimize models just for prediction accuracy. The obtained model may have high computational complexity, so it may not appropriate for embedded or mobile applications with limited resources such as restricted power, low computational resource, and low amount of memory space. Moreover, Complex building-blocks are also hard-to-understand and cannot be engineered later. Optimizing models for multiple objectives (e.g. computational complexity, power consumption, latency, size of network, etc.) can suppress these issues, which is the concept of MONAS.  \cite{hsu2018monas}.

Back to the NAS optimization equation (\autoref{eq:nas2}), MONAS is when $cost(.)$ is a vector of possibly two or more objectives, so the minimization compromises multi-objective (or even, many-objective) optimization. Multi-objective optimization is the heart of MONAS.

The definition of multi-objective optimization is discussed in the following subsection. 

\subsection{Multi-objective optimization} \label{sec:moo}
Multi-objective optimization (MOP) deals with the optimization formula when there is more than one target objective \cite{moosurvey}. Consider two objectives of the accuracy and inference latency on a certain hardware; having a number of possible answers, there will be a trade-off between the two objectives which yields to a range of answers. A solution $a$ may have better accuracy but worse latency than solution $b$ and vice versa. The trade-off exists and will be more complicated when there are more than two objectives. The literature of multi-objective optimization provides solutions and modellings for the problem and discusses the related concepts.

A multi-objective optimization problem can be defines as follows \cite{moosurvey}:

\begin{multline}
\\x^* = \argmin{x}{f(x)}, \\
subject\ to\ g_j(x) <= 0, \qquad j = 1,2,...,m \\
h_l(x) = 0, \qquad l=1,2, ..., e \\
f(x) = [f_1(x), f_2(x), ..., f_k(x)]^T \\
\end{multline}

in which, $k$ is the number of objectives, $m$ is the number of inequality constraints, and $e$ is the number of equality constraints. The space of all possible values for $x$ is called the \textit{design space}, and the \textit{cost} (or \textit{objectives) space} is defined as $\{f(x)|x\in \Omega_x\}$.

The solution of a multi-objective optimization problem is the set of all \textit{non-dominating} solutions (instead of a single optimal solution). A solution $a$ is called \textit{dominating} solution $b$, if $f_i(a) \le f_i(b)$ for $i \in {1, 2, ..., m}$ and $f_i(a) < f_j(b)$ for at least one index $j \in {1, 2, ..., m}$ (and all $f_i$ have to be minimized) \cite{nassurvey2}. The set of all non-dominating solutions is called the \textit{Pareto-optimal} or \textit{Pareto set} which can be visualized as \textit{Pareto front} in the objectives space. The whole Pareto set is considered the solutions of the multi-objective optimization, each of which are a trade-off in the results. In other words, for each $a$ in the Pareto set, there isn't any other answer that is better than $a$ in all the objectives.

In the practical side, we may have various objectives for $f_i$ and even defining and finding good objectives are not trivial. We continue the discussion about the objectives usable for MONAS.

\subsection{The objectives for MONAS}
The Performance Estimation stage of NAS deals with the objective(s) of the networks. Defining and evaluating the objectives is a part of performance estimation in MONAS, as we want to deal with multiple objectives, rather than only the network error on the evaluation set.

Optimizing the network error (or accuracy) is always necessary in any single-objective NAS or MONAS procedure. There may be also interest in minimizing the model size, inference latency, computational cost, robustness of the model, and even the training duration. Any combination of these goals makes a MONAS problem definition.

\autoref{tabobjs} provides an overview of the specifications of different objectives that possibly may be useful for MONAS and the properties of each one. In the following, we elaborate the specifications of the objectives for MONAS.

\begin{table*}[htbp]
\begin{center}
\caption{A list of objectives usable for Multi-objective NAS.}
\label{tabobjs}
\scalebox{0.99}{
\begin{tabular}{l p{1cm} p{1cm} p{2.5cm} l p{3cm}}
\hline
Objective & Notation & Range & Typical Scale & Determinism & Lightness (Heavy/Light) \\
\hline
Inference Accuracy    & $Acc$   &   0..100  & [\%] &   Stoc.  & H \\
Validation Error    & $E_{val}$   &   0..100 & [\%]  &   Stoc.  & H \\
Inference Computational Cost    & $CC_{infer}$   &   0..$\infty$  & [Flops] &  Det.  & L \\
Inference Latency    & $T_{infer}$   &   0..$\infty$  & [m.sec]  & Stoc.  & H \\
Params\# / Memory Footprint Size & $SZ$ & 0..$\infty$ & [MParam], [MBytes] & Det. & L\\
Adversarial Robustness & $E_{val, adv}$ & 0..100 & [\%] & Stoc. & H \\
Robustness to a specific degradation & $E_{val, deg}$ & 0..100 & [\%] & Det. & H\\
Complexity of the architecture (aka. Description Length) & $MDL$ & 0..$\infty$ & [Bits] &  Det. & L\\
Training Duration & $T_{train}$ & 0..$\infty$ & [Minutes] & Stoc. & H \\
\hline
\end{tabular}
}
\end{center}
\end{table*}

\begin{table*}[]
\caption{Number and type of objectives used in recent papers on MONAS}
\label{tab:t1}
\begin{center}
\scalebox{.99}{
\begin{tabular}{|c|l|l|l|l|l|l|l|}
\hline
\multirow{2}{*}{\begin{tabular}[c]{@{}c@{}}\# of \\ objectives\end{tabular}} & \multicolumn{6}{c|}{Objectives} & \multirow{2}{*}{References} \\ 
\cline{2-7}
                                           &                              Accuracy               & Model size             & \begin{tabular}[c]{@{}l@{}}Inference\\ latency\end{tabular} & \multicolumn{1}{c|}{\begin{tabular}[c]{@{}c@{}}Computational \\ Complexity\\ {[}flops{]}\end{tabular}} & Robustness            & Other &                                                                                                                      \\
\hline
\multirow{3}{*}{2} & \multicolumn{1}{c|}{+} & \multicolumn{1}{c|}{+} & \multicolumn{1}{c|}{}                                       & \multicolumn{1}{c|}{}                                                                                  & \multicolumn{1}{c|}{} & \multicolumn{1}{c|}{}                                                                         & \cite{chen2018joint,wu2019multi,yang2020cars,yue2020effective,loni2020deepmaker,huynh2022lightweight}                      \\ 
\cline{2-8} 
                                                                            & \multicolumn{1}{c|}{+} & \multicolumn{1}{c|}{}  & \multicolumn{1}{c|}{+}                                      & \multicolumn{1}{c|}{}                                                                                  & \multicolumn{1}{c|}{} & \multicolumn{1}{c|}{}                                                                         & \cite{tan2019mnasnet,elsken2018efficient,cai2018proxylessnas,schorn2020automated,jiang2020efficient,shi2019multi,he2020milenas,yao2020sm,cheng2020instanas,bosman2021local}                      \\ \cline{2-8} 
                                                                            & \multicolumn{1}{c|}{+} & \multicolumn{1}{c|}{}  & \multicolumn{1}{c|}{}                                       & \multicolumn{1}{c|}{+}                                                                                 & \multicolumn{1}{c|}{} & \multicolumn{1}{c|}{}                                                                         & \cite{lu2019nsga,wang2019evolving,cui2019fast,yao2020sm,fang2020densely,lu2020nsganetv2,zhu2020real,lu2020multiobjective,ma2021scenenet}                     \\ \hline
\multirow{4}{*}{3}                                                          & \multicolumn{1}{c|}{}  & \multicolumn{1}{c|}{+} & \multicolumn{1}{c|}{}                                       & \multicolumn{1}{c|}{+}                                                                                 & \multicolumn{1}{c|}{} & \multicolumn{1}{c|}{PSNR}                                                                     & \cite{chu2020multi,chu2021fast}                      \\ \cline{2-8} 
                                                                            & \multicolumn{1}{c|}{+}                      & \multicolumn{1}{c|}{+}                      &                                                             & \multicolumn{1}{c|}{+}                                                                                                      &                       &                                                                                               & \cite{xiong2019resource}                      \\ \cline{2-8} 
                                                                            & \multicolumn{1}{c|}{+}                      & \multicolumn{1}{c|}{+}                      &                                                             &                                                                                                        & \multicolumn{1}{c|}{+}                     &                                                                                               & \cite{yue2020effective}                      \\ \cline{2-8} 
                                                                            & \multicolumn{1}{c|}{+}                      &                        & \multicolumn{1}{c|}{+}                                                           & \multicolumn{1}{c|}{+}                                                                                                      &                       &                                                                                               & \cite{chen2020multi, zhou2018neural}                      \\ \hline
4                                                                           & \multicolumn{1}{c|}{+}                      & \multicolumn{1}{c|}{+}                      & \multicolumn{1}{c|}{+}                                                           & \multicolumn{1}{c|}{+}                                                                                                      &                       &                                                                                               & \cite{dong2018ppp}                 \\ \hline
5                                                                           & \multicolumn{1}{c|}{+}                      &                        & \multicolumn{1}{c|}{+}                                                          &                                                                                                        &                       & \multicolumn{1}{c|}{\begin{tabular}[c]{@{}c@{}}Hardware-\\ specific\\ objectives\end{tabular}} & \cite{schorn2020automated}                     \\ \hline
\end{tabular}
}
\end{center}
\end{table*}

\begin{figure*}[h]
\begin{center}
\includegraphics[width=.85\linewidth]{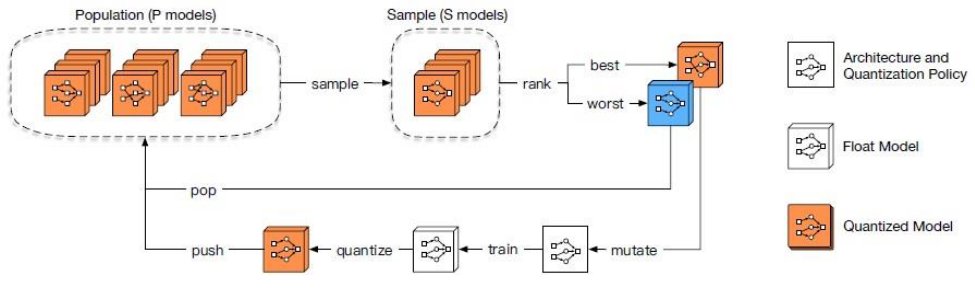}
\end{center}
\caption{Joint neural architecture search and quantization model \cite{chen2018joint}}
\label{fig:JASQ}
\end{figure*}

In \cite{chen2018joint,yang2020cars,wu2019multi,lu2019neural,jiang2020efficient,elsken2018efficient,shi2019multi,loni2020deepmaker,bosman2021local}, authors considered accuracy and model size as two objectives of MONAS. An integrated framework named JASQ was proposed in \cite{chen2018joint}, which incorporated the neural architecture search and quantization technique. It used tournament selection in its evolutionary search method. Also, it used a linear scaling function to quantize the weight vector for given quantization bit. As shown in \autoref{fig:JASQ}, each model in the population is assessed based on accuracy and model size. The best model is quantized again and added to the population. Also, the worst model is removed from population.
In \cite{wu2019multi}, an evolutionary pruning technique was used to lighten the architecture. It prunes the network to achieve two objective: sparse ratio of network and the accuracy. To obtain better result, the pruned network is fine-tuned again. Authors in \cite{jiang2020efficient} proposed a CNN-based search technique named MOPSO/D-Net. It uses multi-objective particle swarm optimization based on decomposition (MOPSO/D) to maximize the accuracy and minimize the number of network parameters, simultaneously.

\begin{figure*}[htbp]
\begin{center}
\includegraphics[width=.85\linewidth]{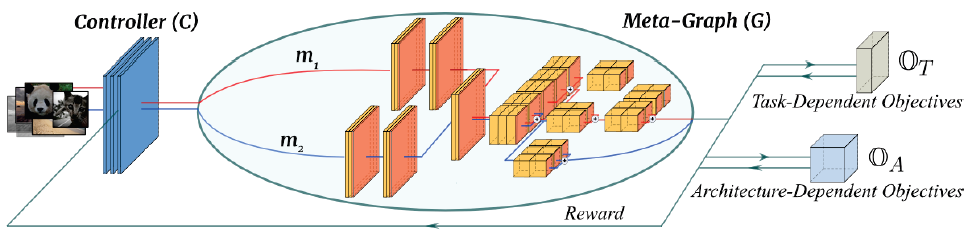}
\end{center}
\caption{InstaNAS:controller choose an expert child architecture (m) from the meta-graph (G), while considering task-dependent objectives such as accuracy and architecture-dependent objectives such as latency. \cite{cheng2020instanas}}
\label{fig:instanas}
\end{figure*}

In \cite{cheng2020instanas,cai2018proxylessnas,tan2019mnasnet}, authors considered accuracy and inference latency as two objectives of MONAS.
A MONAS was proposed in \cite{cheng2020instanas} named InstaNAS, which is the first NAS with instance awareness. As shown in \autoref{fig:instanas}, a controller chooses an expert child architecture with considering the accuracy and latency of architecture. Accuracy of InstaNAS is similar to MobileNet-v2 but in terms of latency it outperforms the MobileNet-v2.

In \cite{hsu2018monas,wang2019evolving,ma2021scenenet,lu2019nsga,fang2020densely,cui2019fast,lu2020nsganetv2,zhu2020real,lu2020multiobjective,yao2020sm}, authors considered accuracy and computational cost as two objectives of MONAS. In \cite{zhang2019neural}, accuracy and hardware efficiency were considered as two objectives of MONAS. In \cite{li2020ftt}, researchers considered accuracy and fault resilience in edge devices as two objectives of MONAS.

Some papers such as \cite{chu2020multi,chu2021fast} used MONAS to handle super-resolution in image domain. Both of these papers used number of parameters, computational cost, and peak signal noise ratio (PSNR) as three objectives of MONAS. In \cite{yue2020effective}, authors proposed a method to find a neural architecture, which is effective, efficient, and robust . They used accuracy, model size, and robustness as three objectives of MONAS. Both \cite{chen2020multi,zhou2018neural} exploited MONAS with three objectives (i.e., accuracy, computational complexity, and inference latency) to find a suitable neural architecture.

We use Many-Objective NAS name for NAS with more than three objectives. In \cite{dong2018ppp}, a many-objective NAS was proposed, which found a architecture with maximum accuracy and minimum model size, computational complexity, and inference latency. Also, a many-objective NAS with five objective was proposed in \cite{schorn2020automated}, which try to optimize the architecture based on accuracy, inference latency, and Hardware-specific objectives, including error resilience, energy efficiency, and Bandwidth requirement. Most important papers on MONAS are summarized in \autoref{tab:t1} with a view to the number and type of objectives.


\subsubsection{Number of objectives used}
The two-objective optimization is the most popular in MONAS \cite{lu2019nsga,hsu2018monas, elsken2018efficient,jiang2020efficient,cheng2020instanas,yao2020sm,chen2018joint,cai2018proxylessnas,wang2019evolving,yang2020cars,tan2019mnasnet,zhang2019neural,wu2019multi,ma2021scenenet,shi2019multi,loni2020deepmaker,fang2020densely,cui2019fast,li2020ftt,lu2020nsganetv2,bosman2021local,zhu2020real,lu2020multiobjective,lu2019neural}, as it is simpler to analysis and is one step forward from a single-objective NAS.
Not surprisingly, the accuracy (or error) is always needed to be optimized, while considering one other goal including latency, model size, etc. As stated in \autoref{fig:taxonomy}, the most usual combinations are accuracy+model size, accuracy+inference latency, accuracy+computational cost, and accuracy+robustness.

Notable work for three or more objectives are \cite{chu2020multi,chu2021fast,xiong2019resource,dong2018ppp,yue2020effective,chen2020multi,schorn2020automated,zhou2018neural}.
It is obvious that more objectives makes the problem harder, may need more tries, and is harder to understand or visualize. Therefore little work is done on three or more objectives for MONAS.

\subsubsection{Stochastic vs. Deterministic Objectives}
Among other reviews on the NAS, we are the first to elaborate the stochastic characteristics of some NAS goals and will show that the stochastic phenomena exists in some objectives, which should be considered.

\begin{figure}[htbp]
\begin{center}
\includegraphics[width=1\linewidth]{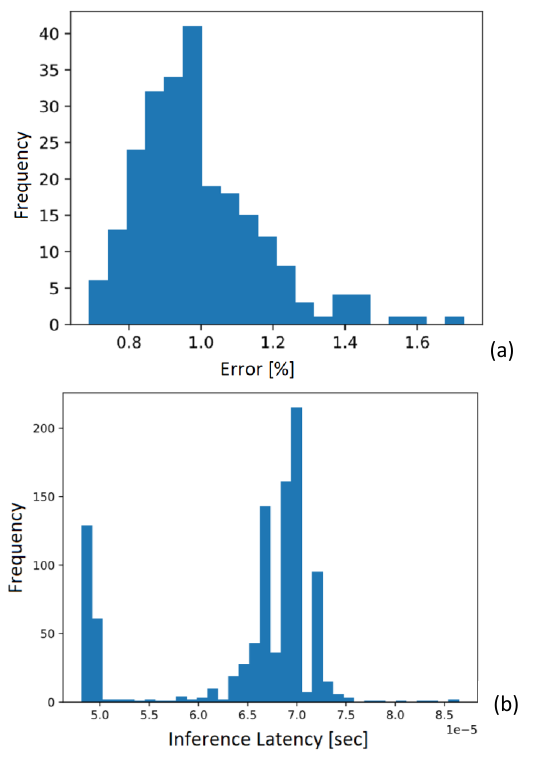}
\end{center}
 \caption{Illustration of the stochastic properties of a single CNN in several evaluations; (a) Training Error, (b) Inference Time.}
\label{fig:stochastic}
\end{figure}

 Some optimization goals like network size, the number of parameters, and computational cost (for a certain architecture) are deterministic properties. On the other hand, some other objectives, especially the accuracy, and the inference latency are stochastic. For any deterministic objective, any trial results in a certain fixed value (for example, number of parameters of network $a_1$). In opposite, each measurement of a stochastic objective differs from the other as it follows a random probability distribution. Consider several trials of training a neural architecture on a fixed dataset from scratch. The value of accuracy may have variations in various trials (and even it may not converge in some trials) based on initial random values and possibly other randomnesses in the learning algorithm. The inference latency is another example. Current state of Clock-rate policy and system thread schedule that is even related to current temperature of the device directly affects the inference latency and may change in each trial. \autoref{fig:stochastic} illustrates the distribution of the error rate as well as inference latency for a single trained architecture and the same input for several trials. 


\subsubsection{Lightweight vs. costly Objectives} \label{sec:light}
Among all objectives that are listed in \autoref{tabobjs}, the computation of some objectives is fast and straight forward, e.g. counting the number of add/multiplication operations, the number of model parameters, while the computation of others is time/resource consuming, e.g. inference latency, training duration. This feature has a side effect on the performance estimation of candidate neural networks and total search time. Some researchers tried to get rid of time/resource consuming objectives and to be able to explore larger search space in a shorter amount of time. The number of parameters in \cite{loni2020deepmaker}, computational complexity in terms of FLOPs in \cite{lu2019nsga}, estimation of cycle count in \cite{achararit2020apnas} are used for simplicity as light weight objectives that are the indicator of network complexity. 

In \cite{benmeziane2021hardware}, the possible ways of obtaining the value of objectives such as latency are categorized in four ways: real-time measurements, lookup table models, analytical estimation and prediction models. Real-time measurement means to measure the latency, as an example objective, on a real device like what MNASNet \cite{tan2019mnasnet} did. Real-time measurements are accurate but considerably time consuming and unscalable for all available devices. The last three ways are used to reduce the computation cost of costly objectives. Lookup table is usually prepared in advance and filled with required metrics to compute the overall latency using operator's latency \cite{wu2019fbnet,abdelfattah2020best}. In \cite{cai2020once}, the inference latency of a network in an implementation scenario (GPU, CPU, mobile deployment) is predicted using a latency lookup table. Analytical estimation computes an approximate estimate using the processing time, the stall time, and the starting time \cite{marchisio2020nascaps}. Prediction models \cite{cai2018proxylessnas,zhang2020fast,nasnet} are ML models that are trained on benchmark’s real-time measurements such as NAS-Bench-101 \cite{ying2019bench} and NAS-Bench-201 \cite{dong2020bench} or self made datasets. The prediction models are not accurate but have better results than analytical estimation and require a large dataset and a considerable preparation time.

Validation accuracy, as listed in \autoref{tabobjs}, is a costly objective and its speed up idea is introduced in \autoref{sec:naspe}. MONAS approaches also use the same idea as single objective NAS approaches to reduce the computation cost of validation accuracy. In \cite{huynh2022lightweight} to avoid long search time, the number of model parameter is used as one of objectives and for another objective an estimation of validation accuracy, suggested in \cite{mellor2021neural}, is used only at initialization phase and when a child network is generate by mutation. This strategy reduces the total search time to find optimal network to 4.3 GPU days.

\subsection{Performance Estimation for MONAS}    
Performance estimation in NAS methods is usually based on the accuracy of candidate architectures/models. It is discussed in \autoref{sec:naspe}, where a categorization of different performance estimation methods is presented. In MONAS researches, performance estimation depends on more than one objective, where the accuracy usually exists in the set of objectives. Therefore, similar methods have been used in MONAS methods. \textcolor{red}{for example.} In \cite{dong2018ppp} an RNN regressor is trained to estimate the validation accuracy of candidate network architectures. For other objectives, the estimation is discussed in \autoref{sec:light}.

The overall performance in MONAS is estimated based on the following two methods, i.e. scalarization or Pareto frontier.
 
\subsubsection{MONAS by Scalarization}
By scalarization, a scalar value is obtained from a vector of multiple objectives. One of the most common scalarization methods is to use the weighted sum of objectives. Weighted sum could be done by fixed weights or by random weights that is called random scalarisation.

For the first time, a MONAS framework was proposed in \cite{hsu2018monas} for multiple objectives neural architecture search, which was built on top of the NAS with reinforcement learning \cite{zoph2016neural}. After the MONAS framework, more papers have been published in the field of MONAS that have improved the architecture based on different objectives. The recent studies can be classified by different aspects of the NAS such as number of objectives, search strategies, search space, application, architecture type, and architecture features. A taxonomy for MONAS is illustrated in \autoref{fig:taxonomy}.

\begin{figure}[h]
\begin{center}
\includegraphics[width=.95\linewidth]{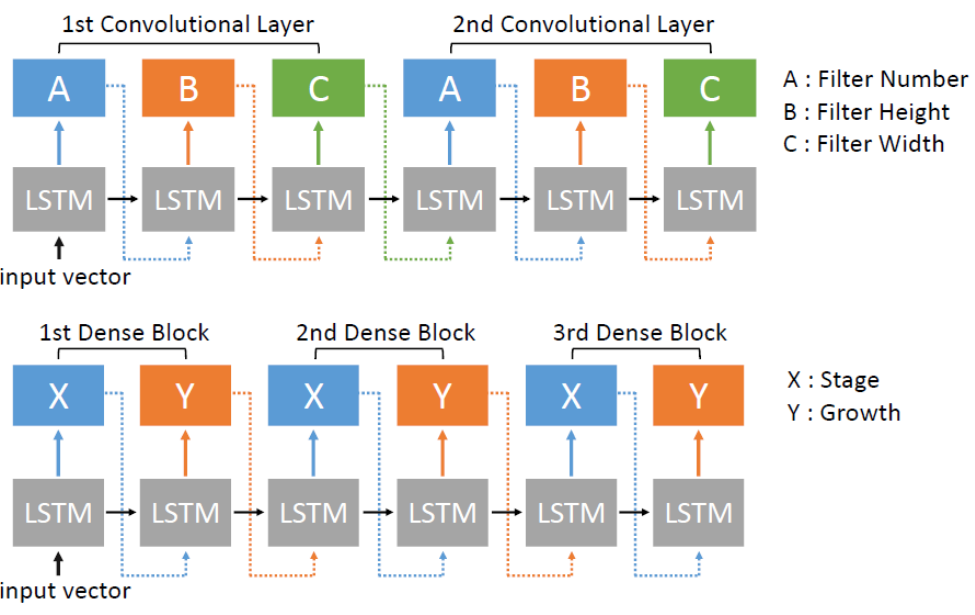}
\end{center}
 \caption{RNN workflow for AlexNet and CondenseNet \cite{hsu2018monas}}
\label{fig:RNNworkflow}
\end{figure}

MONAS \cite{hsu2018monas} has two stage, including generation stage and evaluation stage. In first stage, a RNN was used as a robot network (RN) to generate a sequence of hyperparameters in a CNN. \autoref{fig:RNNworkflow} shows the RN, an RNN structure with one-layer LSTM. In second stage, the CNN model was trained as a target network (TN) with the hyperparameters of first stage. The accuracy and power consumption of the trained TN are rewarded to the RN. The reward function is a weighted sum of accuracy and power consumption. Then, RN was updated using these rewards with reinforcement learning. 

Gradient-based MONAS \cite{cai2018proxylessnas}, like MONAS \cite{hsu2018monas}, uses the scalarization method to make the gradient-based optimization algorithm be possible.
However, different researchers use different formulas for scalarization, sometimes other than weighted sum, e.g., ProxylessNas \cite{cai2018proxylessnas} used $ACC(m). [LAT(m)/T]^\lambda$ as the optimization goal where $ACC(m)$ denotes the accuracy of model m, $LAT(m)$ denotes the latency of model m, $T$ is the target latency and $\lambda$ is the hyperparameter of the trade-off between accuracy and latency.

\subsubsection{MONAS without scalarization, Pareto Frontier}
\begin{figure}[h]
\begin{center}
\includegraphics[width=.95\linewidth]{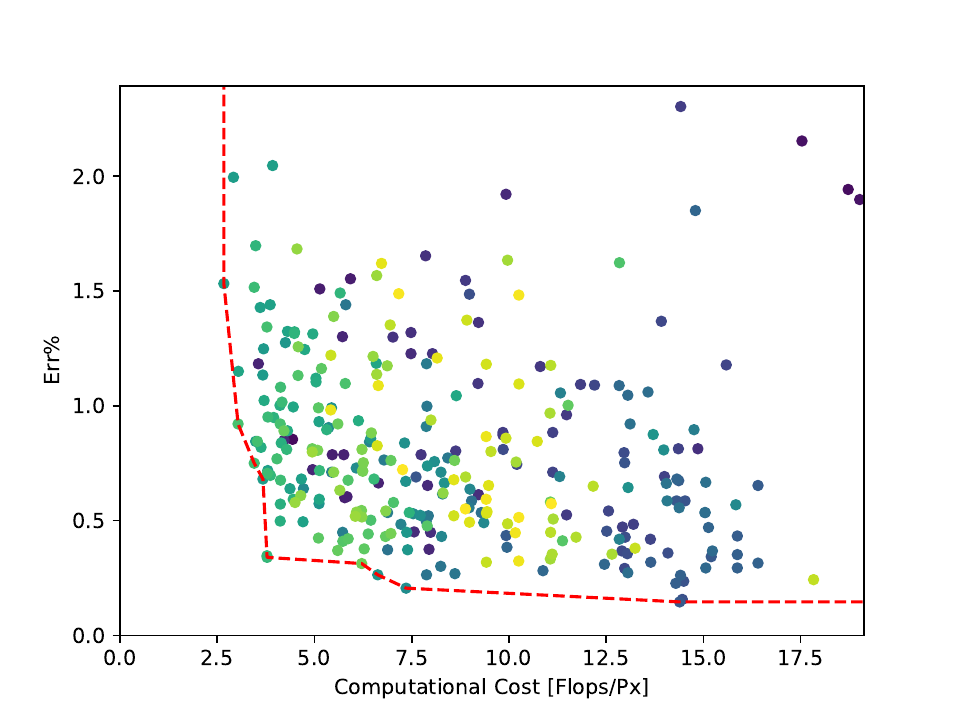}
\end{center}
 \caption{The Pareto Frontier of a Multi-Objective Optimization Solution.}
\label{fig:pareto}
\end{figure}

As it is discussed in \autoref{sec:moo}, for multi-objective problems a solutions \textit{a} is called dominating solution \textit{b}, if \textit{a} is equal or better than \textit{b} in all objectives and better for at least one objective. A solution is said to be Pareto optimal if and only if any other solution to the problem does not dominate it. The set of all Pareto optimal solutions is called Pareto frontier. MONAS approaches that do not use of scalarization try to find a good approximation of Pareto frontier by means of algorithms like NSGA-II \cite{deb2002fast}. \cite{lu2019nsga, chu2020multi, loni2020deepmaker} use NSGA-II algorithm to find the best architectures in the search space. In \autoref{fig:pareto}, a Pareto frontier of a two-objective optimization problem is shown in its objective space.

\subsection{Search Methods for MONAS}
Although most researches used reinforcement learning (RL) and evolutionary methods as search strategies, some researches applied other methods such as Bayesian optimization, stochastic search, and gradient-based. The most important papers are categorized in \autoref{tab:t2} according to the different search strategies. Search strategies refer to different methods for exploring the search space. 

NAS can use reinforcement learning to train an agent that can learn a search policy. The performance of the model (i.e. evaluated in each iteration) is used in order to reward the agent and select an architecture that leads to higher performance. Finally, a model with a maximum performance will be picked out.

Evolutionary methods consider a population of neural networks. In process of evolution, a network from the population is selected as a parent and generates children using mutations on them. Mutations in NAS are carried out using manipulation in network such as adding a shortcut connection, adding or removing a layer, altering the hyper-parameters. Then the performance of the child is estimated and they was accumulated in population.

\begin{table}[]
\begin{center}
\caption{papers on different search strategies}
\label{tab:t2}
\begin{tabular}{ll}
\hline
Search Method & Reference         \\ \hline
RL              & \cite{cheng2020instanas,hsu2018monas,cai2018proxylessnas,tan2019mnasnet,zhang2019neural,cui2019fast,li2020ftt,chen2020multi,lu2019neural,zhou2018neural} \\
EA              & \cite{chen2018joint,wang2019evolving,yang2020cars,wu2019multi,ma2021scenenet,lu2019nsga,loni2020deepmaker,yao2020sm,lu2020nsganetv2,zhu2020real,lu2020multiobjective,schorn2020automated}   \\
RL + EA         & \cite{chu2020multi,chu2021fast}              \\
Local Search    & \cite{bosman2021local}              \\
RNN             & \cite{dong2018ppp}              \\  
Bayesian        & \cite{shi2019multi}              \\ 
other           & \cite{xiong2019resource}              \\ 
\hline
\end{tabular}
\end{center}
\end{table}

\subsection{Search Space Definitions in MONAS}
In contrast to the performance estimation and the search method, there is no difference between the search space of a single-objective NAS with multiple-objective ones. This is because the search space has nothing to do with the outputs or objectives of the NAS. However, we will have a look to the search spaces used in MONAS research.

To find the best architecture, the search space including different architectures should be explored. The size of the search space can be too large so the search process can be very time consuming. The size of the search space can be diminished using knowledge about characteristic of different architectures. Although it can reduce the search time it also lead to prevent finding novel solutions that is beyond our knowledge. In \autoref{tab:t_spacesize}, we listed some examples of search space size.

\begin{table}[]
\begin{center}
\caption{Some examples of search space size. The size of the search space can be too large so the search process can be very time consuming.}
\label{tab:t_spacesize}
\begin{tabular}{ll}
\hline
Method name                       & Search space size                                                                          \\ \hline
InstaNAS \cite{cheng2020instanas} & 10\textasciicircum{}25                                                                     \\
MOPSO/D \cite{jiang2020efficient}                          & 2\textasciicircum{}12, 2\textasciicircum{}18, 2\textasciicircum{}25, 2\textasciicircum{}33 \\
LEMONADE \cite{elsken2018efficient}                          & 300                                                                                        \\
MONAS \cite{hsu2018monas}                            & 1.6 * 10\textasciicircum{}29                                                               \\
NSGA-Net \cite{lu2019nsga}                          & 1200                                                                                       \\
MoreMNAS \cite{chu2020multi}                         & 192*7                                                                                      \\
JASQNet/ JASQNet-Small \cite{chen2018joint}           & 3.5*10\textasciicircum{}9                                                                  \\
MnasNet \cite{tan2019mnasnet}                          & 10\textasciicircum{}39                                                                     \\ \hline
\end{tabular}
\end{center}
\end{table}

\begin{figure}[h]
\begin{center}
\includegraphics[width=0.80\linewidth]{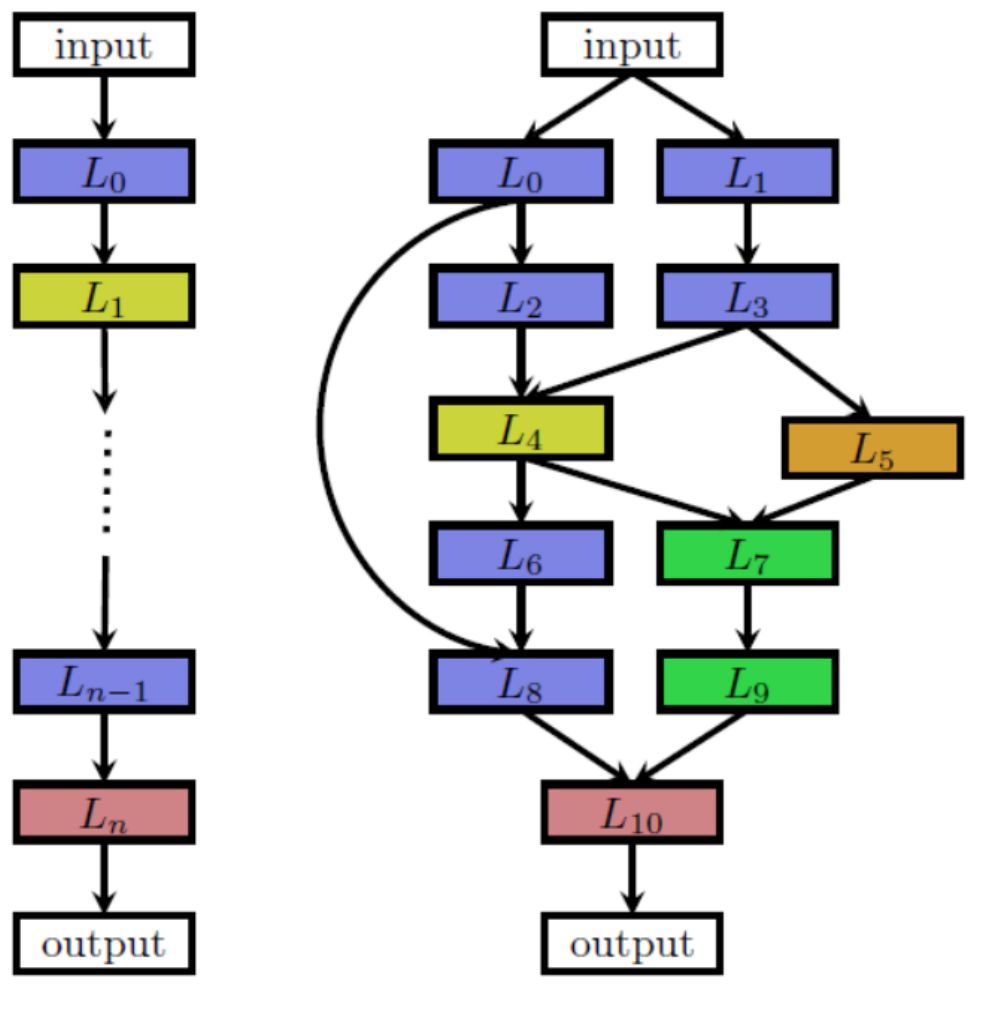}
\end{center}
 \caption{Different architecture spaces, including chain structures and multibranch structures \cite{nassurvey}.}
\label{fig:archspace1}
\end{figure}

Architectures in the search space can have different structures of neural network, including chain structures and multibranch structures \cite{nassurvey}, which is illustrated in \autoref{fig:archspace1}. The chain structure is a neural network with sequential layers, whereas multibranch structure is a more complex architecture with multiple branches such as Residual Networks and DenseNets.

\begin{figure}[h]
\begin{center}
\includegraphics[width=0.95\linewidth]{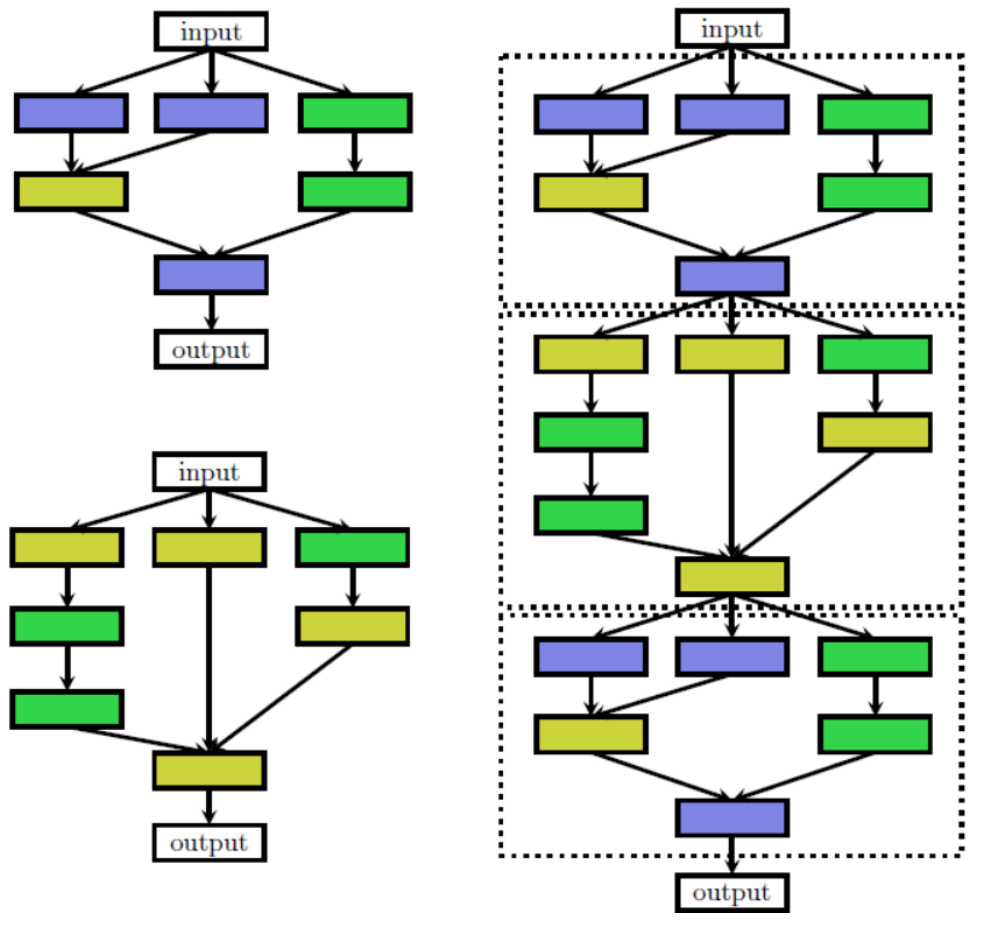}
\end{center}
 \caption{(Left) two different blocks, including normal block and reduction block. (Right) a neural architecture by stacking the blocks \cite{nassurvey}.}
\label{fig:blocksearch}
\end{figure}

In another aspect, NAS can find the appropriate architecture by considering the whole architecture (e.g. \cite{hsu2018monas} or just concentrate on block optimization (e.g. \cite{jiang2020efficient}). Also, some works consider both approaches to find an appropriate architecture (e.g. \cite{lu2020multiobjective,chen2020multi,zhu2020real,bosman2021local,yue2020effective,yao2020sm,li2020ftt,cui2019fast,fang2020densely,loni2020deepmaker,lu2019nsga,dong2018ppp,ma2021scenenet,cheng2020instanas}). Two different blocks, including normal block and reduction block are illustrated in \autoref{fig:blocksearch} (Left). Also, an architecture is illustrated in \autoref{fig:blocksearch} (Right) by stacking these blocks. 

\cite{fernandez2020searching} proposes to use Winogard convolutions that are the fastest known algorithm for spatially small convolutions. It presents wiNAS as a framework that can find the optimal convolution algorithm among im2row or different Winograd implementation and consider both high accuracy and low latency. It decreases the latency as well as the performance estimation time.

Weight sharing in One-Shot NAS is an approach to reduce the search time/cost. However, this approach does not guaranty a good accuracy in application. Progressive Automatic Design of search space is proposed in \cite{xia2022progressive} to add diversity in the search space. The authors use a variation of NSGA-II to handle multi-objective approach and balance between latency and accuracy.

Pasca \cite{zhang2022pasca} is a NAS system for GNNs with an emphasis on scalability and applicability for web-scale graphs. Balancing between accuracy and inference time via multi-objective optimization is another contribution of this article that is related to the topic of our paper. The search space of \cite{zhang2022pasca} and its underlying application is different from common applications in the field of NAS/MONAS. 

\subsection{by application}
MONAS can be used in wide range of applications such as image and video processing, object detection, image classification, semantic segmentation \cite{LIU2021104195}, object tracking \cite{DBLP:journals/corr/abs-2104-14545b}, super-resolution, and natural language processing. MONAS approaches are categorized based on application in \autoref{tab:app}.

LiDNAS is proposed for lightweigth monocular depth estimation task in \cite{huynh2022lightweight}. It is a multi-objective approach and its objective function is a weighted sum of validation accuracy and the number of model parameters.

\begin{table*}[]
\begin{center}
\caption{papers on different applications of MONAS}
\label{tab:app}
\begin{tabular}{ll}
\hline
Application      & Reference \\ \hline
Image Classification & \cite{cheng2020instanas,jiang2020efficient,elsken2018efficient,hsu2018monas,chen2018joint,cai2018proxylessnas,xiong2019resource} \\
 & \cite{wang2019evolving,yang2020cars,tan2019mnasnet,zhang2019neural,ma2021scenenet,lu2019nsga,loni2020deepmaker,fang2020densely,cui2019fast,li2020ftt,yue2020effective,lu2020nsganetv2,bosman2021local,zhu2020real,chen2020multi,lu2020multiobjective} \\ 
 & \cite{fernandez2020searching,schorn2020automated,lu2019neural,dong2018ppp} \\ 
Object Detection     & \cite{yao2020sm,yao2020sm,tan2019mnasnet}    \\
Image Segmentation   & \cite{zhang2019neural,lu2019nsga}         \\
Super-Resolution     & \cite{chu2020multi,chu2021fast}         \\ 
Depth estimation & \cite{huynh2022lightweight} \\ 
Disease detection & \cite{miahi2022genetic} \\
Remote sensing scene classification & \cite{ma2021scenenet} \\
\hline
\end{tabular}
\end{center}
\end{table*}

\section{Other Practical Aspects in MONAS}

\subsection{Software Packages} 
Some researchers have published the codes or generated models of their proposed NAS/MONAS approach \cite{real2019regularized,lu2019nsga,chu2021fast,xiong2019resource,chu2020multi,tan2019mnasnet,lu2020nsganetv2}. Pymoo package (https://pymoo.org/) in some of these codes, like \cite{lu2019nsga}, is the basis of coding. Pymoo is a software package in python and offers state of the art single/multi objective optimization algorithms and many more features such as visualization. It is widely used and referenced by MONAS papers. \cite{lu2019nsga} also uses and modifies DARTS codes \cite{darts}. Moreover, BenchENAS \cite{xie2022benchenas} is a platform in that nine state-of-the-art ENAS algorithms are implemented and it is straight forward to conduct fair comparisons between evolutionary algorithm based NAS methods. Also, \cite{bestpractices} enumerates lists of best practices for releasing code as well as comparing NAS methods and discusses the properties of a good open source library for NAS methods which is also applicable for MONAS.

\subsection{Datasets Used}
Although any dataset can be used for MONAS domain, some datasets have received more attention in this area. For example CIFAR-10 \cite{krizhevsky2009learning}, CIFAR-100 \cite{krizhevsky2009learning}, ImageNet \cite{deng2009imagenet} and COCO \cite{lin2014microsoft} have been mostly used in the area of MONAS researches.

Canadian Institute For Advanced Research presented a dataset CIFAR-10, which contain 60,000 32x32 color images. CIFAR-10 is widely used in machine learning applications such as object detection and classification. Images of this  dataset labeled in 10 different classes (each class contains 6,000 images), including trucks, ships, cars, airplanes, birds, deer, cats, frogs, dogs, and horses.

CIFAR-100 dataset is similar to the CIFAR-10, but it contains 100 classes (each class contains 600 images). Each class divided into two groups, including 500 images for training and 100 images for testing. The CIFAR-100 has 20 super-classes. Each image has two labels: class label and super-class label. 

The most common datasets that used in MONAS are listed in \autoref{tab:datasets}.

\begin{table*}[]
\begin{center}
\caption{Common datasets in the literature of MONAS}
\label{tab:datasets}
\begin{tabular}{ll}
\hline
Dataset                                          & Used in \\ \hline
CIFAR-10 \cite{krizhevsky2009learning}            & single objective: \cite{nasnet,real2019regularized}, multi-objective: \cite{lu2019nsga,lu2020nsganetv2,fernandez2020searching,elsken2018efficient,dong2018dpp,cai2018proxylessnas,loni2020deepmaker,dong2018ppp,yue2020effective,chen2020multi} \\
CIFAR-100 \cite{krizhevsky2009learning}           & multi-objective: \cite{lu2019nsga,loni2020deepmaker,lu2020nsganetv2,fernandez2020searching,elsken2018efficient,xiong2019resource,yue2020effective,chen2020multi} \\ 
ImageNet \cite{deng2009imagenet}                 & single objective: \cite{nasnet,real2019regularized}, multi-objective:  \cite{howard2019searching,lu2020nsganetv2,tan2019mnasnet,cai2018proxylessnas,tan2019mnasnet,xiong2019resource,chen2020multi} \\
MNIST \cite{lecun1998mnist}                     & multi-objective: \cite{loni2020deepmaker} \\
COCO \cite{lin2014microsoft}                     & single objective: \cite{nasnet}, multi-objective: \cite{tan2019mnasnet} \\
Others\textsuperscript{*} &  multi-objective: \cite{huynh2022lightweight,miahi2022genetic,chu2020multi,chu2021fast,yue2020effective} \\
\hline
\multicolumn{2}{l}{* Others are NYUDepth-v2 \cite{silberman2012indoor}, KITTI \cite{Geiger2012CVPR}, ScanNet \cite{dai2017scannet}, MHSMA \cite{javadi2019novel} datasets, \etal} \\ 
\hline
\end{tabular}
\end{center}
\end{table*}

\subsection{Hardware resource need} 
In this section, a list of resources that are used in different researches is provided in \autoref{tab:hwr}. Note that these resources are consumed in different conditions, applications and datasets. Therefore, they are not directly comparable. They only give us a sense of the amount of resources needed.

Moreover, in order to have a fair comparison between evolutionary based NAS method, a platform is developed in \cite{xie2022benchenas} with several ENAS algorithms. It is possible to run algorithms in the same environment and with the same settings and compare them in terms of accuracy, flops, parameter size and GPU days.

\begin{table*}[]
\begin{center}
\caption{Resources used in the literature of MONAS}
\label{tab:hwr}
\begin{tabular}{ll}
\hline
Reference                                          & Hardware description \\ \hline
\cite{nasnet} & 2000 GPU hours (single objective), Nvidia P100s \\
\cite{real2019regularized,real2017large} & 3150 GPU days (single objective), k40, P100 \\
\cite{huynh2022lightweight} & 4.3 GPU days \\
\cite{miahi2022genetic} & 240 GPU hours \\
\cite{lu2019nsga} & 4 or 8 GPU days \\
\cite{lu2020nsganetv2} & 1 GPU days \\
\cite{loni2020deepmaker} & 75 GPU hours \\
\cite{tan2019mnasnet} & 4.5 TPUv2 days \\
\cite{chu2020multi} & 56 GPU days \\
\cite{chu2021fast} & 24 GPU days \\
\cite{xiong2019resource} & 4 or 16 GPU days \\
\cite{yue2020effective} & 0.836 GPU day \\
\cite{chen2020multi} & 75 or 60 GPU hours + 99 additional GPU hours to pre-train the shared model \\ 
\hline
\multicolumn{2}{l}{Note that most of reported days/hours are approximate values and} \\
\multicolumn{2}{l}{they are related to different GPU models and} \\
\multicolumn{2}{l}{even on different datasets.}\\
\hline
\end{tabular}
\end{center}
\end{table*}

\section{Future Directions}
In this section, important issues that are interesting to research will be introduced.

Participation in the development of related software packages and platforms is a possible future work. For example, BenchENAS which is a platform for comparing EA based NAS methods, welcomes participation to contribute it. It is useful to continue the way of works that try to provide a unified evaluation platform for popular NAS/MONAS methods \cite{ying2019bench,dong2020bench}.

Based on \autoref{tab:t1}, most papers use two or three objectives in their performance estimation of candidate architectures and many-objective NAS researches are very limited. Researching in this direction and combining different conflicting objectives can be useful. Besides combining different objectives, defining new objectives could be another future work. As far as we know, complexity of the model based on minimum description length (MDL) is not used in any other paper as an objective. It is a string representation of neural network based on description length of its building blocks. Defining a cost on model complexity helps NAS to find compact and non-complex neural networks. In this paper, complexity of the model and robustness to a specific degradation are proposed as new objectives. Researchers could use these new objectives and other new objectives in their MONAS researches. These objectives could be application specific, e.g. model robustness with respect to blur in image processing. Such robustness related objectives have direct positive effect on validation error and inference accuracy.

A large amount of previous work was related to reduce the computational cost of the overal NAS procedure. This even lead to different methods such as one-shot NAS. The computational cost reduction is yet a direction of NAS research. Research on performance prediction, predictors for objectives like validation accuracy, latency, memory peak usage, etc., could be useful to have an efficient performance estimation. It can be done in general or for specific type of settings such as specific architectures.

Moving towards application is a useful direction for future researches. Based on \autoref{tab:app}, current applications are more focused on image classification using CNN based architectures . This line can be extended to other applications and new usage areas. More mature applications than classification on CIFAR-10 or CIFAR-100, e.g. medical diagnostic applications, driver assistance applications, text to speech, speech to text and language models in NLP related applications are some common applications. As CNN based architectures are not enough for speech/text/other applications where LSTM, transformers and etc. are more popular, this direction can establish another direction to research about other building blocks in DNN architectures. In previous surveys, attention mechanism and transformer layers have not been expanded yet. Building blocks such as transformers, vision transformers \cite{dosovitskiy2020image,khan2021transformers,liu2021swin}, \etal could be added to the search space of NAS/MONAS methods. If we want to use deep learning models on edge a more diverse set of hardware must be reviewed and considered. This issue restricts the applicability of previous researches for other researchers that want to use the best architecture for a specific hardware. It seems that more diverse search spaces, datasets, tasks, and hardware must be included in the future researches.

Beside known building blocks, identify most useful building blocks and inferring DNN design patterns is another direction to guide other researchers for selecting useful building blocks based on application or other important factors. Transfer learning is a common technique in deep learning and it is of interest in NAS/MONAS approaches to fine-tune weights for datasets other than training dataset. In MONAS, it is more challenging because of the presence of objectives other than accuracy like latency, power consumption, \etal. Transfer learning is appropriate to transfer learned architectures from small dataset to larger dataset to initialize the search space properly, decrease the search time and improve the scalability. There are proposed architecture sets and it is good to extend them for a diverse set of applications.

Explainability shows how significant each of model parameters contribute to the final decision. It helps users to trust a deep learning model, mitigates risks, detect potential biases, characterize model accuracy, fairness, transparency and outcomes, and ensure that the system is working as expected. It is a trend nowadays and MONAS researchers could use the optimality provided by explainability.

RL based NAS are very time/resource consuming search methods, on the other hand, GD search methods requires expert knowledge about designing a good super-net. Evolutionary NAS needs less time/resource to search and has little dependence on expert knowledge. Apparently, this is the reason for the excessive spread of ENAS methods. Why not leave the design of a differentiable super net to evolutionary algorithms to take the advantages of GD search methods?

It is difficult to compare different NAS/MONAS approaches in a fair manner. Different approaches have different search algorithms, search spaces, performance estimation strategies, available software/hardware, \etal. Although \cite{xie2022benchenas} has addressed the issues of fair comparisons and efficient evaluations for ENAS, this problem is still open and needs more research for ENAS and other approaches (RL, GD, BO, \etal). Which parameters have more effect on architecture, how they affect on architecture, many other questions that can be answered by post analysis of applied search methods and their final results.

\section{Conclusion}
In this survey, we have provided a comprehensive study of Multi-Objective Neural Architecture Search (MONAS). We presented a well-defined formulation for NAS and corrected classification mistakes found in previous surveys. We also provided a detailed list of used and possible objectives for MONAS, which included considerations for computational complexity, power consumption, and network size. Furthermore, we discussed the stochastic nature of objectives in NAS, such as model accuracy, in the context of multi-objective optimization, which is often treated as a deterministic measure.

Overall, our review sheds light on the current state-of-the-art in MONAS and highlights several important research directions for the future. We believe that the continued development of NAS techniques will lead to more efficient and effective neural network architectures that can meet the demands of increasingly complex machine learning tasks. Additionally, exploring novel objectives and optimization algorithms for MONAS can help researchers better understand the trade-offs between different objectives and improve the efficiency and robustness of the search process.

\scriptsize
\bibliographystyle{IEEEtran}
\bibliography{bibs}

\end{document}